# Automatic Color Image Segmentation Using a Square Elemental Region-Based Seeded Region Growing and Merging Method

Hisashi Shimodaira

**Abstract**—This paper presents an efficient automatic color image segmentation method using a seeded region growing and merging method based on square elemental regions. Our segmentation method consists of the three steps: generating seed regions, merging the regions, and applying a pixel-wise boundary determination algorithm to the resultant polygonal regions. The major features of our method are as follows: the use of square elemental regions instead of pixels as the processing unit, a seed generation method based on enhanced gradient values, a seed region growing method exploiting local gradient values, a region merging method using a similarity measure including a homogeneity distance based on Tsallis entropy, and a termination condition of region merging using an estimated desired number of regions. Using square regions as the processing unit substantially reduces the time complexity of the algorithm and makes the performance stable. The experimental results show that our method exhibits stable performance for a variety of natural images, including heavily textured areas, and produces good segmentation results using the same parameter values. The results of our method are fairly comparable to, and in some respects better than, those of existing algorithms.

## 1 INTRODUCTION

Image segmentation is a process in which all the pixels of an image are classified into a number of regions. It is desired that the resultant regions correspond to constituent objects or their parts in the image. The regions are used in subsequent processes including image understanding and object recognition. Various techniques for producing a good image segmentation have been proposed in the literature. The existing image segmentation techniques are roughly classified into three categories: feature-space based techniques, image-domain based techniques, and physics based techniques [1]. Image-domain based techniques include region growing approaches. The method proposed in this paper belongs to the seeded region growing (SRG) approach subset of the region growing approaches.

Here we consider what a good image segmentation should be. The conditions of a good image segmentation listed in [2] are as follows. (a) Regions should be uniform or homogeneous with respect to some characteristics such as color tone or texture. (b) Region interiors should be simple and without many holes. (c) Adjacent regions should have significantly different values with respect to the characteristics on which they are uniform or homogeneous. (d) Boundaries of each region should be simple, not ragged, and must be spatially accurate. In natural images, the distributions of characteristics such as color tone or texture are complex and vary considerably between images. Achieving all of these conditions in natural image segmentation is difficult. Therefore, developing a better segmentation method has been one of the challenges in computer vision.

Existing SRG methods are classified into two approaches: the concurrent region growing approach [3], which our method belongs to, and the recursive region growing approach [4]. We shall briefly review the problems of the former SRG approach and existing methods within this approach.

A. Seed generation

SRG starts with seeds assigned manually or automatically. We deal with the automatic seed generation approach. To produce accurate segmentation results, the generated seeds must satisfy the following three conditions. First, the seed pixels must be located in uniform or homogeneous areas. Second, for a desired region, at least one seed must be generated to produce this region [5]. Third, seeds for different regions must be disconnected [5]. Each connected component of seed pixels is considered one seed region and assigned a respective label (region number). To reduce the computational complexity of the seed growing process, the number of the seed regions should be as small as possible [6].

Existing methods generate the seed pixels by using the following methods: the J value and the semi-adaptive threshold value [7], centroids between adjacent color edges [8], the H-value and the semi-adaptive threshold value [9], the similarity measures computed by color components in a 3 × 3 pixel area and the predefined threshold values [5], centroids of rectangles [10], color gradient and the adaptive threshold generation [11], and the minimum between the fuzzy similarity and non-edge fuzzy membership and the adaptive threshold values [12].

B. Seed Region growing

---

★ The author has no affiliations.
  Email: hshimodaira@c06.itscom.net

After seed generation, the seed regions are grown by incorporating a pixel among the unlabeled pixels into an adjacent seed region. The image is segmented into plural regions by repeating this process until all of the image pixels have been labeled. Because region boundaries are formed at the location where plural regions meet, the criterion and order for incorporating pixels into adjacent seed regions are critical for locating accurate region boundaries. Because the SRG process is inherently sequential, the final segmentation results depend on the order in which pixels are processed [13]. Such order dependency occurs whenever several pixels have the same difference measure to their adjacent regions.

Regarding the criterion, the following are used: the city block distance [3], [8] or the relative Euclidean distance [5] between a pixel and the adjacent region in color space. In [12], the product of the fuzzy distance between a pixel and the adjacent region and the fuzzy edge membership degree of the pixel was used. Regarding the order for incorporating pixels into adjacent seed regions, the ascending order of the criteria mentioned above are used.

C. Region boundary determination

In the seed region growing process, when a pixel meets plural adjacent seed regions, the decision criterion by which the appropriate seed region is selected is critical for locating accurate region boundaries.

Regarding this problem, the criteria mentioned above are also used except in [12], where the fuzzy distance between a pixel and the region is used.

D. Region merging

Because most of the existing seed region growing methods result in over-segmented regions, the regions are merged in subsequent processes. The region merging process is a type of combinatorial optimization process. Therefore, the merge criterion and merge order directly affect the quality and coarseness of the segmentation result.

Regarding the merge criterion, most of the SRG methods use the Euclidean distance between the mean color components of two adjacent regions as in [5]. Others use the following criteria: the Euclidean distance between color histograms [7], the L1 distance of color histograms [9], the fuzzy distance between mean color components [12], and the multivariate analysis [11]. Regarding the merge order, all the methods use the ascending order of the merge criteria. Regarding the termination criterion, most of the methods use a predefined threshold value.

SRG has a potential to allow us to implement a good segmentation algorithm. However, existing SRG methods have the following weaknesses.
* Regarding the seed generation, the similarity measures or the gradient values of color components are, with a few exceptions [7], [9], computed using a small window (e.g., 3 × 3 pixels). Most characteristics of texture in real images cannot be captured by such a small number of pixels. This leads to the generation of inappropriate seeds or the failure to generate appropriate seeds.
* The ability to accurately locate region boundaries is weak, because in the seed region growing processes, edge or gradient information is not used except in [12]. A method of integrating edge (contour) and region information is required to locate accurate boundaries [14], [15].
* For the termination criteria, a predefined threshold value is used. This causes the problem described below.

Most real-world images contain textured areas. Various methods of texture analysis have been proposed [16], [17]. However, their appropriateness for color image segmentation is limited by the factor including the high computational complexity and the estimation of texture model parameters. How to deal with the combination of color and texture remains an open and challenging problem.

Most existing image segmentation methods have several parameters that must be predefined by the user. Their values strongly affect the segmentation quality and coarseness. Because their optimum values depend on the image characteristics, they are different for every image. Consequently, the user must perform a time-consuming trial–and-error task to tune them for every image under study. This is one of the problems requiring a solution [18].

Against these backgrounds, we propose an automatic image segmentation method using square elemental region-based seeded region growing and merging (S-SRG) method, which belongs to the concurrent region growing approach. Our goal is to provide the user with an empirically general segmentation consisting of significant segments corresponding to not only salient objects, but also to the surroundings. The resultant segmentation is intended to be used in subsequent processes such as image understanding and object recognition. The method is intended to be used on a natural image.

The following are features and innovations of our method:
* An image segmentation method using square elemental regions instead of pixels as the processing unit.
* An automatic seed generation method using detection of the local minima of enhanced gradient values.
* A seed region growing method using a growth control oriented distance and boundary localization oriented distance.
* A region merging method using a similarity measure including a homogeneity distance based on Tsallis entropy and a merge termination condition using an estimated desired number of regions of the segmentation result.
* A mutually most similar region merging method for reducing over-segmentation.

The remainder of this paper is organized as follows. Section 2 introduces the outline of our method. Section 3 presents the preliminary process. Section 4 presents the automatic seed region generation method. Section 5 presents the region merging method. Section 6 presents the pixel-wise boundary determination method. Section 7 shows applied examples of our method. Section 8 shows the experimental results and discusses them. Finally, Section 9 presents the conclusions.

2   OUTLINE OF OUR METHOD

The outline of our method is as follows.

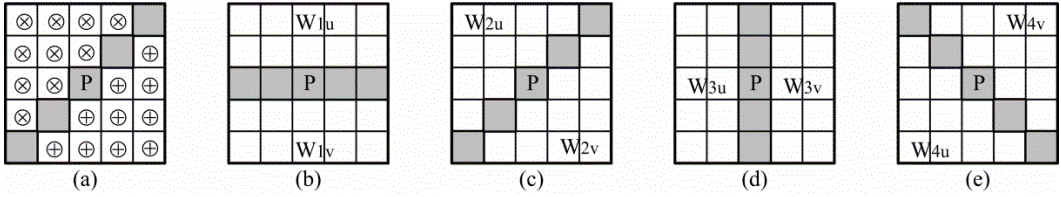

Fig. 1. Observation window for computing the gradient *G*.

(1) Preliminary process
    Transform the color image from the RGB color space to the L*a*b* color space and apply the bilateral filter to the L*, a*, and b* components. Then, compute the magnitudes of color vectors, the salient edges, the Tsallis entropy, and the desired number of regions.
(2) Automatic seed region generation process
    (a) Apply the initial seed region detection algorithm to the color vector magnitude space and obtain the initial seed regions (SRs).
    (b) Apply the seed region growing algorithm to the SRs and obtain the initial polygonal regions (INPRs).
(3) Region-merging process
    Apply region merging methods 1, 2, and 3 successively to the INPRs. We refer to the resultant regions as the polygonal region 1s (PR1s), the polygonal region 2s (PR2s), and the final polygonal regions (FPRs), respectively.
(4) Pixel-wise boundary determination process
    Apply the pixel-wise boundary determination algorithm to he FPRs and obtain the final segmentation result. Finally, the boundaries of the FPRs are refined by locating the accurate region boundaries through this process.

In our method, the processes from generating the SRs through to computing the FPRs are performed using small square elemental regions (SERs), which are formed by dividing the image in a non-overlapping way. We refer to our method as the square elemental region-based seeded region growing and merging (S-SRG) method. Using square regions instead of pixels as the processing unit implicitly reduces the influence of pixel value variations and unnecessary details. This makes the performance stable and robust. Pixel-based region growing methods are computationally intensive [19]. Using square regions substantially reduces the time complexity of the algorithm.

Regarding the geometrical neighborhood (adjacency) relationship of a pixel (SER), we use the four-neighbor relationship. This means that if a pixel (SER) has at least one neighboring pixel (SER) belonging to a region, the pixel (SER) is adjacent to the region.

Our method has several tunable parameters. The parameters with a direct relation to each image characteristic have different optimal values for every image. To solve this problem, we normalize the measure representing each image characteristic by the global maximum and minimum values. Such a device allows us to use the same parameter values for every image. The optimal parameter values determined by the experiments are described at the related Sections. With these values, the user can obtain nearly optimal results for almost every natural image without tuning.

## 3 PRELIMINARY PROCESSES

### 3.1 Color Space

A color image is usually specified in the RGB color space. However, the distance in the RGB color space does not represent the perceptual difference in a uniform scale. We use the CIE L*a*b* color space to represent color features. It is a perceptually uniform color space and the Euclidean distance in it corresponds to the color difference in human perception.

In the preliminary process, the color space of the image is first transformed from the RGB space to the L*a*b* space. Then, the combined bilateral filtering [20] is applied to the *L*\*, *a*\*, and *b*\* components. We apply this filter twice, using a sliding 7 × 7 pixel window and the parameters $\sigma_d = 2$ and $\sigma_r = 8$. This has the effect of averaging only perceptually similar colors and preserving only perceptually important edges [20]. Expectedly, compared to the case without filtering, we obtained better segmentation results by applying this filtering.

We compute gradient and entropy values of the image using the monochrome image obtained by computing the magnitude of the color vector (*CV*) at each pixel using (1) and normalizing it into [0, 255].

$$CV = \sqrt{L^{*2} + a^{*2} + b^{*2}}. \qquad (1)$$

This value can represent the characteristics of a color image better than the lightness correlate *L*\*.

### 3.2 Salient Edge Extraction Method

In the seed region generation process, we use edge information as in [8], [12]. To this end, we must detect salient edges which can be the boundaries of salient objects or object parts. Although filters based on a 3 × 3 pixel window are used in [8], [12], they cannot detect salient edges in images containing textured regions. This is because texture is a local-neighborhood property and such a small size filter cannot capture the texture characteristics, and thus discriminate between different textures. To detect salient edges, edges must be computed at a scale that is appropriate for capturing the characteristics of the underlying texture structure.

To achieve this goal, we propose a local histogram difference based salient edge extraction (LHDSEE) method as follows. The window W of a $L_W \times L_W$ pixel area centered on the pixel under study is set as shown in Fig. 1a. The four pairs of the symmetric sub-windows, $W_{kU}$ and $W_{kV}$, for $k = 1, ..., 4$, are defined in the window W, as shown in Figs. 1b, 1c, 1d, and 1e. Such a window scheme is used in [21]. Four pairs are used as a compromise between the angular resolution of edge directions and the computational complexity. The $CV$ values observed in each sub-window form the histograms $U_k$ and $V_k$, for $W_{kU}$ and $W_{kV}$, respectively. Each histogram is normalized and the histogram difference $G_k$ is computed as

$$G_k = \sum_{i=1}^{M} | U_{ki} - V_{ki} |, \quad (2)$$

where $M$ is the number of the histogram bin, and $U_{ki}$ and $V_{ki}$ are the histogram components of $U_k$ and $V_k$, respectively. The bin width is determined using the maximum and minimum values of the $CV$ values and $M$. Then, the gradient $G$ at the pixel is computed as

$$G = \max_{k=1,\ldots,4} G_k. \quad (3)$$

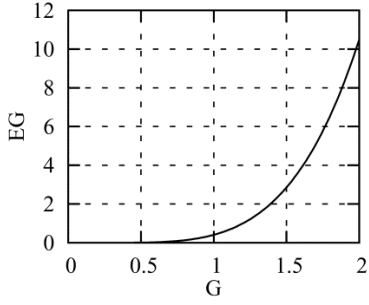

Fig. 2. The relation between the gradient $G$ and the enhanced gradient $EG$ for $\delta = 0.2$.

Then, the enhanced gradient $EG$ is computed as

$$EG = (G - \delta)^{\gamma}, \quad \text{for } \delta \leq G \leq 2,$$
$$EG = 0, \quad \text{for } 0 \leq G \leq \delta, \quad (4)$$

where $\delta$ is the offset and $\gamma$ is the magnifying coefficient. The gradient $G$ has a value in [0, 2]. Fig. 2 shows the relation between the $G$ and $EG$ for $\delta = 0.2$. The maximum value of $EG$ is 10.4976.

We use $L_W = 11$, $M = 20$, $\delta = 0.2$, and $\gamma = 4.0$. By this function, weak edges are neglected or further weakened, and strong edges are emphasized.

### 3.3 Tsallis Entropy

In computing the desired number of regions and the similarity measure for the region merging 2, we use the Tsallis entropy [22]. It is defined as

$$S_q = \frac{1}{q-1} \left\{ 1 - \sum_{i=1}^{M} p_i^q \right\}, \quad (5)$$

where $p_i$ denotes the $i$th component of the normalized histogram of the values under study and $M$ the number of the histogram bin. The bin width is determined using the global maximum and minimum values of the values under study and $M$. We use $q = 0.8$, and $M = 20$.

In computing the desired number of regions, we compute the $S_q$ values at each pixel using an 11 × 11 pixel sliding window on the $CV$ image. In computing the similarity measure for the region merging 2, we compute them for each region using the color components of pixels in the region.

### 3.4 Desired Number of Regions

In image segmentation, the produced regions should be significant and consist of not only salient objects or salient object parts but also the surroundings in the image. In addition, the produced regions should satisfy the conditions of good image segmentation listed in Section 1. Therefore, we attempt to estimate the desired number of such regions and use it as the criterion to determine the termination condition in the region merging 2 process. We assume that the desired number of regions is a function of the image complexity: the higher the image complexity, the larger the desired number of regions. We use the fractal dimension of the image to estimate the image complexity.

The fractal dimension of a surface corresponds quite closely to our intuitive notion of roughness [23]. The fractal dimension of an image takes a value between 2 and 3 according to the characteristics to the image. The higher the proportion of rough areas in the image, the higher the fractal dimension will be. Such behavior is appropriate for estimating the image complexity for our purposes.

However, because the fractal dimension of a natural image is strongly influenced by the proportion of textured area, it is not appropriate for our purposes. We empirically found that the average ($F_A$) of the fractal dimensions of the $EG$ image ($F_{EG}$) and that of the Tsallis entropy image ($F_{TE}$) is an appropriate measure of image complexity for this method. We define the desired number of regions $N_D$ as

$$N_D = \alpha (F_A - 2.0)^{\kappa}, \quad (6)$$

where $\alpha$ and $\kappa$ are adjustable parameters. We compute the fractal dimension using the method in [24]. We determined the appropriate values of $\alpha = 170$ and $\kappa = 2.0$ according to the results of the proposed segmentation method.

## 4 AUTOMATIC SEED REGION GENERATION METHOD

### 4.1 Dividing Image into SERs

First, we divide the image into non-overlapping SERs of side length of $S$. Because the experimental results showed that the optimal value of $S$ is 4 pixels, we use $S = 4$ pixels.

### 4.2 Initial Seed Region Detection

The generated seed regions must satisfy the three conditions listed in Section 1. First, the initial seeds must be located in uniform or homogeneous areas. We use the $EG$ values to detect initial seeds fulfilling this condition. In the $EG$ image, $EG$ values at the boundaries between different uniform regions or different homogeneous texture regions are high, whereas those inside these areas are low. Therefore, we determine initial seeds by detecting local minima in the $EG$ values and areas neighboring them.

The process is as follows.
(1) Compute the average of the $EG$ values of pixels belonging to each SER and let it be $EG_a$.
(2) Compute the average of the $EG$ values of pixels belonging to all the SERs and let it be $EG_A$.

(3) Compute the threshold value $T_S$ as
$$T_S = \beta\, EG_A. \quad (7)$$
(4) Classify the SERs: if the $EG_a$ value is less than or equal to $T_S$, then let the SER be a Seed SER (SSER); otherwise, let the SER be a Non-Seed SER (NSSER).
(5) Compute the connected components of the SSERs based on the four-neighbor connectivity and assign a label (region number) to each connected component.

Each connected component of the SSERs is considered one seed region. Therefore, seed regions can be one SER or one region having several SERs. The coefficient $\beta$ is a predefined parameter. The value of $\beta$ must be determined so as to satisfy the second and third conditions for seed region generation listed in Section 1. If it is too small, the second condition cannot be satisfied. If it is too large, the third condition cannot be satisfied. We empirically found its optimal value to be 0.4 and therefore use $\beta = 0.4$.

### 4.3 Seed Region Growing

In this process, all of the NSSERs are in turn classified into the most similar region among the adjacent seed regions and the initial polygonal regions (INPRs) are formed. The initial seed regions are located in areas that are approximately uniform or homogeneous in image characteristics. Therefore, the resultant regions become as uniform or homogeneous as possible. The criteria for classification and processing order used in the existing methods have the problems mentioned in Section 1. Therefore, we developed an algorithm for addressing these weaknesses. It differs from the conventional algorithms [3], [5] in its use of growth control oriented distance (*GCD*) and boundary localization oriented distance (*BLD*).

The similarity between two regions is measured by the distance between the regions: the more similar they are, the smaller the distance is. We use a different distance for each purpose. The basic distance is the Euclidean distance *CD* between points 1 ($L^*_1, a^*_1, b^*_1$) and 2 ($L^*_2, a^*_2, b^*_2$) in the color space defined as follows.
$$CD = \sqrt{(L^*_1 - L^*_2)^2 + (a^*_1 - a^*_2)^2 + (b^*_1 - b^*_2)^2}. \quad (8)$$
Computing the distance between SERs or regions, we use the average of the color components of the pixels belonging to each SER or region.

We define the *GCD* between an NSSER and the adjacent seed region as in (9) and perform the seed growing processing in the ascending order of *GCD* values.
$$GCD = CD + \omega EG_a, \quad (9)$$
where *CD* is computed using (8), $EG_a$ is the average of the *EG* values in the NSSER, and $\omega$ is a predefined parameter.

Deciding which seed region to assign an NSSER to, we use the *BLD* defined as follows.
$$BLD = CD + \lambda_1 G_S, \quad (10)$$
where *CD* is computed using (8), $\lambda_1$ is a predefined parameter, and $G_S$ is the *CD* between the NSSER and the adjacent SER belonging to the seed region. If there are plural adjacent SERs, the average of the $G_S$ values is taken.

To store the pairs of NSSER numbers and *GCD* values, we use a sequential sorted list (SSL) [3]. In a preliminary process, the following processing is performed: search for an NSSER adjacent to at least one initial seed region, compute the *GCD* value, and store them in the SSL; after processing all the relevant NSSERs, sort the data in the SSL in the ascending order of *GCD* values. If an NSSER has plural seed regions adjacent to it, the minimum value of the *GCD* values is taken. Because the *GCD* values of different NSSERs never have the same value, the order dependency mentioned in Section 1 never occurs in our method.

The algorithm consists of the following three steps repeated while the SSL is not empty.
(1) Remove the first NSSER, X, from the SSL and check the labels of its four neighbors. Then, we have the following three cases. First, if only one neighbor is labeled, assign X this label. Second, if plural neighbors are labeled and the labels are the same, assign X this label. Third, if plural neighbors are labeled and the labels are different, compute the *BLD* values between X and the adjacent seed regions and assign X the label of the seed region having the minimum *BLD* value.
(2) Update the color component averages of the seed region R to which X is added.
(3) Check the four neighbors of X and select the NSSERs of Y that are not yet stored in the SSL. Then, compute the *GCD* values between Y and the seed region R and add the data to the SSL in the ascending order of the *GCD* values.

In our algorithm, the approximate region boundary is formed by two adjacent SERs whose labels are different from each other (boundary SERs). One of them should be located on the true boundary. Our algorithm works to effectively form appropriate seed regions to achieve this in the following ways.

⋆ If an NSSER is on a salient edge, its *GCD* takes a larger value than NSSERs not on salient edges due to the second term $EG_a$ in (9). As a result, the NSSER on the salient edge is processed later than NSSERs not on salient edges. This leads to increasing possibility that relevant boundary SERs meet on salient edges.

⋆ The second term $G_S$ in (10) is the approximate gradient between the two relevant SERs. Therefore, using (10) in the third case of the algorithm (step (1)) increases the ability to select the most appropriate region. As a result, this makes it possible to accurately locating the boundary pixels in the pixel-wise boundary determination.

We set the parameter values of $\omega$ and $\lambda_1$ to the empirically optimal values of $\omega = 5.0$ and $\lambda_1 = 2.0$.

## 5 REGION MERGING METHOD
### 5.1 Region Merging Strategy

Up until now, elaborate region merging methods have been proposed: e.g., the merge importance based method [25], the merge likelihood based method [26], the Markov random field based method [6], and the overall coding length minimization method [17].

The over-segmented regions produced in the seed region generation process are merged in this process. The resultant regions should be significant and correspond to constituent objects or their parts in the image. Because the global optimization method requires a large amount of computation time, we use a greedy optimization algorithm based on empirical knowledge. Less significant regions are merged into the neighboring region having the minimum value of measures based on empirical knowledge through the region merging 1, 2, and 3 processes.

### 5.2 Region Merging 1

In this process, small regions in the INPRs are merged into their most similar adjacent regions and the polygonal region 1s (PR1s) are produced. This process is required to eliminate small regions that are not appropriate for obtaining the reliable histograms in the region merge 2 process.

The region adjacency relationship is represented by the region adjacency table [10]. The similarity is measured by the $CD$ between regions computed using (8). The maximum size of region ($M_R$) that will be processed is predefined by the number of SERs that constitute the region.

First, we build the region adjacency table and make the list of the regions sorted in the ascending order of region sizes. We then merge a region with size less than or equal to $M_R$ into the most similar adjacent region. This is repeated in the ascending order of region sizes, updating the region adjacency table and the list of the regions after each iteration.

We set the parameter $M_R$ to 5. Therefore, the minimum region size after the region merging 1 process is equal to or more than 80 pixels in the case of SER side length of 4 pixels.

### 5.3 Region Merging 2

In this process, less significant regions in the PR1s are merged into more significant regions and the polygonal region 2s (PR2s) are produced. The processing is performed in the ascending order of the merge importance defined by region size and region similarity. Because real-world images consist of color and texture, a region similarity measure for representing color and texture features is required. We use the following region similarity measure combining the color distance and homogeneity distance.

The color distance $D_A$ between two regions is defined as the average of the three color distances $D$ values for the $L^*$, $a^*$, and $b^*$ components. Each color distance $D$ is defined as [27]

$$D = \sqrt{1-\rho}, \quad (11)$$

and

$$\rho = \sum_{i=1}^{M} \sqrt{p_i\, q_i}, \quad (12)$$

where $\rho$ is the Bhattacharyya coefficient between the normalized vectors **p** and **q**. These vectors are made by the color component histograms of the two regions, where $M$ is the number of the bin. The bin width is determined using the global maximum and minimum values of each color component.

The homogeneity distance $HD$ between two regions is defined as

$$HD = |E_1 - E_2|, \quad (13)$$

where $E_1$ and $E_2$ are the Tsallis entropy values of the two regions, respectively. The Tsallis entropy values of each region is computed as the average of the Tsallis entropy values for the $L^*$, $a^*$, and $b^*$ of each region. The homogeneity distance supports the color distance in discriminating the texture features in the case where the color distance cannot discriminate different textured regions.

The similarity between two adjacent regions is defined as

$$RD = D_A + \xi\, HD, \quad (14)$$

where $\xi$ is a predefined parameter.

The merge importance ($MI$) between two adjacent regions [25] is defined as

$$MI = N_S \times RD, \quad (15)$$

where $N_S$ is the number of SERs constituting the smaller one of the two regions.

The merge processing is performed in the ascending order of $MI$. This is based on the following empirical knowledge. If a small region is very similar to an adjacent region, there is a high possibility that they constitute one region corresponding to a constituent object or object part in the image. Therefore, first, the two regions should be merged. Such merging introduces the smallest change in the segmented image.

The termination condition is defined by the merge importance ratio ($MIR$) [25] as follows.

$$MIR = \frac{MI_{next}}{MI_{max}} > T_t, \quad (16)$$

where $MI_{next}$ is the $MI$ value of the next merge and $MI_{max}$ is the maximum $MI$ value of all preceding merges from the time when this condition is applied. The merging process is terminated, if the $MIR$ value becomes larger than the predefined threshold $T_t$. If this termination condition is applied at an inappropriate time, it results in premature termination. In [25], to prevent this, this termination condition was not applied for the first 10% of all possible merges.

According to our experiments, the timing in applying this termination condition and the value of $T_t$ strongly affected the number of regions in the segmentation results. Moreover, their appropriate values were different for every image. Therefore, we used the estimated desired number of regions $N_D$ as the timing to apply this termination condition: it is applied when the number of regions reaches the $N_D$ value. In this way, we can obtain satisfactory segmentation results using the same value of $T_t$ for every image.

The merge processing is repeated in the ascending order of the $MI$ values until the termination condition is satisfied. The region adjacency table and the region data related to the current merging are updated each time.

We experimentally determined the optimal values of $\xi = 0.1$ and $T_t = 1.04$.

### 5.4 Region Merging 3

In this process, some of the regions in the PR2s are merged based on a mutual similarity measure and the final polygonal regions (FPRs) are produced. This process is based on the following empirical knowledge. Among the regions produced in region merging 2, several pairs of mutually most similar regions exist: they have a distance between them that is smallest among distances between the other adjacent regions. There is a possibility that they are produced, because some salient object or object part is divided into two regions, as a result that different seeds were generated. The smaller the sum of the region sizes is, the higher the possibility is.

In this process, the region similarity is measured by $RD$ defined in (14). The process is as follows. Search and detect the pairs of mutually most similar regions. Then, select the pair with the smallest sum of the region sizes and merge them. The process is repeated, updating the region adjacency table and the region data. The merging process is terminated when the number of the regions reaches the final region number $N_F$ defined as

$$N_F = \varsigma\, N_2 \,, \quad (17)$$

where $N_2$ is the total number of the regions produced in the region merging 2, and $\varsigma$ is a predefined parameter.

We experimentally determined the optimal value of $\varsigma = 0.8$. This is the limit value that does not bring about undesired merging of salient regions for almost all natural images.

## 6 PIXEL-WISE BOUNDARY DETERMINATION METHOD

In this process, the pixel-wise boundary determination algorithm is applied to the FPRs and the final segmentation result is produced. Each pixel of the image is assigned the label of the region (region number) to which it belongs. The region boundary is formed by two adjacent pixels whose labels are different from each other (boundary pixels).

In the segmentation result consisting of the FPRs, the temporary region boundaries are formed by the boundary SERs belonging to different regions. The true boundary is supposed to be located inside one of the boundary SERs. To detect the true boundaries, we repeatedly move the temporary boundary pixels to the supposed true boundary as follows. First, we change the present boundary pixels into free pixels (unlabeled pixels) and then classify them into the most similar region among the adjacent regions using the same algorithm as in the seed region growing described in Section 4.3.

Instead of (9) and (10), we use the distance between a pixel and the adjacent region ($PRD$) defined as

$$PRD = CD + \lambda_2\, G_p \,, \quad (18)$$

where $CD$ is the color distance between the pixel and the adjacent region computed using (8), and $\lambda_2$ is a predefined parameter. The term $G_p$ is the $CD$ between the pixel and an adjacent pixel belonging to the region and computed using (8). If there are plural adjacent pixels, the average of the $G_p$ values is taken. Because $G_p$ is the gradient between the two relevant pixels, the $PRD$ works to accurately locating the boundary pixels as described in Section 4.3.

The following process is repeated a predefined number, $M_P$, of times.
(1) Change the present boundary pixels into free pixels.
(2) Compute the $PRD$ values between all free pixels and their adjacent regions and store the pairs of pixel numbers and $PRD$ values in the SSL. Then, sort the data in the SSL in the ascending order of $PRD$ values.
(3) While the SSL is not empty, repeat the classification processing of the free pixels in the same way as in steps (1) and (2) in Section 4.3.

We experimentally determined the optimal values of $\lambda_2 = 2.0$ and $M_P = 2$ for the case of SER side length of 4 pixels.

## 7 APPLIED EXAMPLES OF THE PROPOSED METHOD

### 7.1 Examples of Processing Results

In this section, we applied the proposed method to two images to demonstrate the result of each process. We used the parameter values described in the preceding text.

Fig. 3 shows the images under study: the flower garden and butterfly.

Fig. 4a shows the $CV$ image for the flower garden.

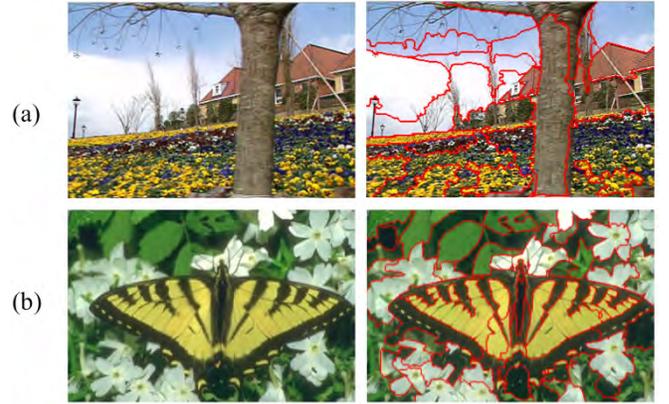

Fig. 3. The original images (left) and final segmentation results (right) of (a) flower garden ($N_F = 23$) and (b) butterfly ($N_F = 54$).

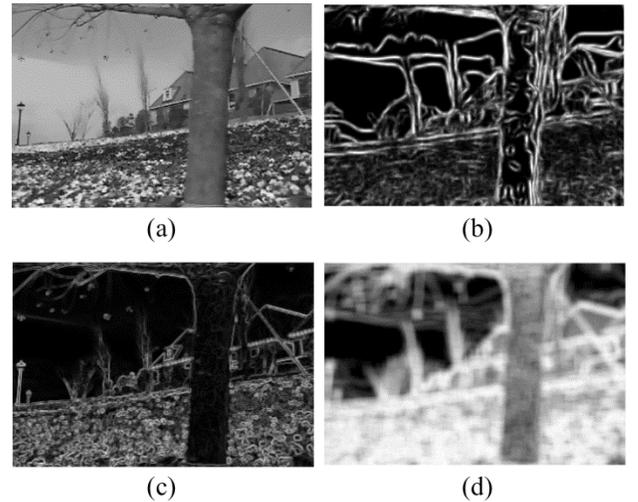

Fig. 4. The results produced by the preliminary process for the flower garden image: (a) $CV$ value image, (b) $EG$ value image, (c) gradient image produced by the Sobel filter, and (d) Tsallis entropy image.

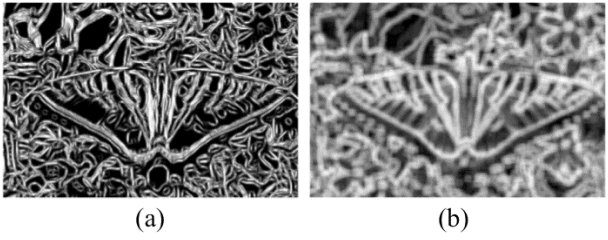

(a)                  (b)

Fig. 5. The results produced by the preliminary process for the butterfly image: (a) *EG* value image and (b) Tsallis entropy image.

TABLE 1. The Fractal Dimension values of the *EG* Image ($F_{EG}$) and Tsallis Entropy Image ($F_{TE}$), and the Desired Number of Regions ($N_D$) for the Flower Garden and Butterfly Images

| Image | $F_{EG}$ | $F_{TE}$ | $N_D$ |
|---|---|---|---|
| flower garden | 2.580 | 2.332 | 35 |
| butterfly | 2.840 | 2.579 | 86 |

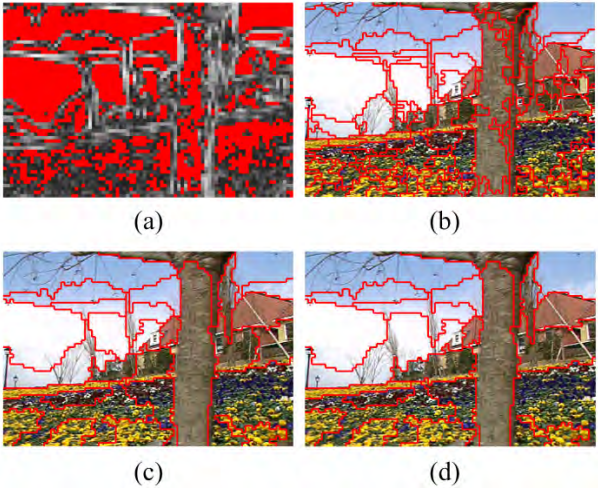

(a)                  (b)

(c)                  (d)

Fig. 6. The results produced in the seed region generation and region merging processes for the flower garden image: (a) detected seed regions (SRs), (b) INPRs produced by the seed region growing, (c) PR2s produced by the region merging 2, and (d) FPRs produced by the region merging 3.

Figs. 4b and 5a show the *EG* images for the flower garden and butterfly, respectively. Fig. 4c shows the image of the gradient values of the flower garden computed by the Sobel filter. These images were made by normalizing the *EG* and gradient values into [0, 255]. Note the following: in the *EG* images, the edges corresponding to boundaries of different texture areas are enhanced, whereas the edges inside each texture area are suppressed. In contrast, in the gradient image produced by the Sobel filter, detecting texture boundaries is difficult, because edges corresponding to constituent parts of the texture are strong. In [21], various methods for detecting texture boundaries were proposed. We tested these, but found the results to be inferior to those of our method.

Figs. 4d and 5b show the Tsallis entropy images. They were made by normalizing the Tsallis entropy values into [0, 255].

Note that the entropy image exhibits very little difference in entropy values within the uniform or homogeneous areas. Table 1 shows the fractal dimension values and $N_D$ computed by (6) using these values.

Fig. 6a shows the detected initial seed regions (SRs) colored with red against a background image of the average *EG* values of the SERs. Note that the proposed initial seed region detection algorithm works well to generate seed regions satisfying the three conditions listed in Section 1. Fig 6b shows the INPR image with boundary pixels colored with red. The total number of the INPRs is 153. Fig 6c shows the PR2 image. The total number of the PR2s is 29. Fig 6d shows the FPR image. The total number of the final regions ($N_F$) is 23. Note that the over-segmented regions generated by the seed region generation process are merged appropriately through the region merging processes, and significant regions are produced.

Fig. 3a shows the image of the final segmentation result ($N_F$ = 23) of the flower garden with the boundary pixels colored with red. Note that the boundaries of the salient objects such as the building roofs are accurately located. The texture boundaries in the flower garden also seem to be appropriately detected and located. Fig. 3b also shows the image of the final segmentation result ($N_F$ = 54) of the butterfly with the boundary pixels colored with red. Note that the boundaries of salient objects are accurately located. The number of regions in the final segmentation of each image seems to be appropriate according to the image complexity of each image.

### 7.2 Effectiveness of the Proposed Methods

In this section, we demonstrate the effectiveness of the proposed region merging methods using experiments in which we change the computation conditions.

First, we changed the value of $\varsigma$ in (17), leaving the values of the other parameters unchanged. Figs. 7a and 7b show the results for $\varsigma = 1.0$ and 0.7 in which $N_F = 29$ and 20, respectively. Comparing Fig. 7a with Fig. 3a, we find that the region merging method 3 works well to produce better segmentation results. However, as shown in Fig. 7b, the over-merging results in undesirable merging that a part of the tree trunk and the surroundings are merged.

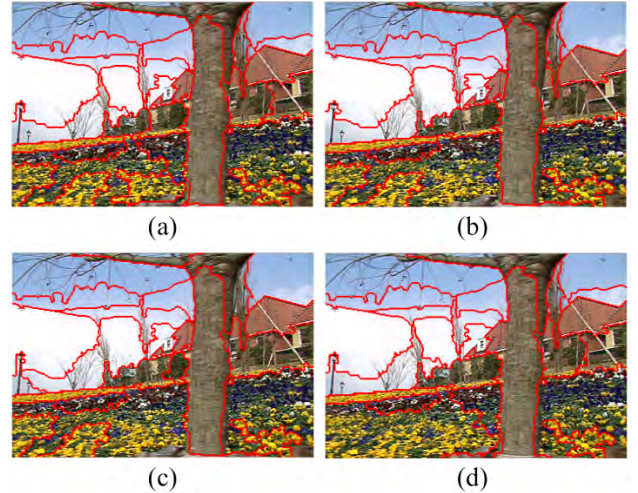

(a)                  (b)

(c)                  (d)

Fig. 7. The results of the experiments on changing the computation conditions: (a) $\varsigma = 1.0$ in (17) ($N_F = 29$), (b) $\varsigma = 0.7$ in (17) ($N_F = 20$), (c) not using the homogeneity distance *HD* in (14) ($N_F = 23$), and (d) not using the region merging 3 process ($N_F = 23$).

Second, we performed the experiment not using the second term (the homogeneity distance $HD$) in (14), leaving the other parameters unchanged. Fig. 7c shows this result. Note that undesirable merging occurs in the results: some parts of the flower garden, and part of the tree trunk and the surroundings are merged. Comparing Fig. 7c with Fig. 3a, we find that the homogeneity distance $HD$ effectively works to discriminate between the different texture regions.

Third, we performed the experiment not using the region merging 3 process in which the final total region number was adjusted to be 23. Fig. 7d shows this result. Note that undesirable merging also occurs in the results: some parts of the flower garden, and part of the tree trunk and the surroundings are again merged. Comparing Fig. 7d with Fig. 3a, we find that the region merging 3 process works well to produce better segmentation results.

## 8 EXPERIMENTAL RESULTS AND DISCUSSIONS

To verify the performance of the proposed method, we performed the experiments on many natural images: standard test images available on the Internet, the images in the Berkeley segmentation database [28], and those in the Stanford dataset [29]. Furthermore, we compared the results produced by the proposed method (S-SRG) with those produced by the existing publicly available algorithms: mean shift (MS) algorithm [30] and graph-based image segmentation (GIS) algorithm [31]. By these means, we investigated features of each algorithm, including their strengths and weaknesses.

For our method, we used the parameter values described in the preceding sections through all experiments. The existing algorithms mentioned above are known to be sensitive to changes in parameter values [18]. However, from the standpoint of fairness, we also used the same parameter values in these algorithms through all experiments. For the MS algorithm, after referring to results shown in [30], [32], we used $(h_s, h_r) = (16, 19)$ and $(17, 19)$ for the small and large (512 × 512) images, respectively, and the minimum region area of 80. For the GIS algorithm, we used the values recommended by the author of [31]: $sigma = 0.5$, $k = 500$, and the minimum region size = 80.

### 8.1 Experimental Results and Visual Comparison

In this section, we select and show the segmentation results of 12 typical images having textured and uniform area characteristics that are as varied as possible. The image names and sizes are as follows: flower garden (352 × 240), rock (481 × 321), horses (481 × 321), leopard (481 × 321), flowers (481 × 321), butterfly (481 × 321), building (320 × 265), coast (321 × 481), boat (321 × 481), baboon (512 × 512), Lenna (512 × 512), and peppers (512 × 512). Figs. 8 and 9 show the original images and the images segmented by the three algorithms. Boundary pixels in S-SRG segmented images are colored with white or red. Table 2 shows the number of regions ($N_F$) in each of the segmented images.

Comparing these results visually in terms of the four conditions of good image segmentation [2] listed in Section 1, we observe the following.

⋆ The numbers of regions in the segmentation results demonstrate that our region merging processes using the estimated desired number of regions works well overall to produce the appropriate number of regions. Compared with our method, GIS, and especially MS, produce significantly higher numbers of regions including many small regions.

⋆ In most of the segmentation results of our method, the significant regions are well separated, whereas in those of MS and GIS, there are many cases in which the significant regions are merged with each other: e.g., the tree trunk and surroundings in Fig. 8a MS, the rock and field in Fig. 8b GIS, parts of the two horses in Fig. 8c MS and GIS, the legs of the horse and the grass in Fig. 8c GIS, part of the leopard and surroundings in Fig. 8d GIS, part of the boat and sea in Fig. 9b GIS, and some of the peppers in 9e GIS. MS completely failed to segment the leopard as shown in Fig. 8d MS.

⋆ In the segmentation results of our method, the boundaries are as a whole spatially accurate and not very ragged, whereas in those of MS and GIS, there are many cases in which the boundaries are ragged, e.g., the tree trunk in Fig. 8a MS and GIS, and the leopard in Fig. 8d GIS. These show that in cases in which the contrast between an object and the surrounding is low, the ability of these two algorithms to accurately locate the boundary is weak.

⋆ For the building image in Fig. 8g, the results of MS and GIS are good compared with the result of S-SRG, because the detailed parts of the objects are accurately segmented. However, if we use a slightly larger desired number of regions $N_D$, our method can produce much better results.

⋆ Overall, the results of our method exhibit relatively stable segmentation qualities, whereas those of MS and GIS are considerably different by image characteristics.

⋆ For the flower garden image, Fig. 8a S-SRG is almost comparable to Fig. 6g in [7]. Fig. 10b in [5] shows that the method proposed in that paper could not separate the tree trunk and the surroundings.

From the results mentioned above, we conclude the following regarding our method.
(1) The seed region generation method works well to generate seed regions that satisfy the three conditions listed in Section 1, as demonstrated by the final segmentation results.
(2) The regions generated by the seed region generation process are appropriately merged through the three region merging processes using the estimated desired number of regions. As a result, an appropriate number of significant regions based on the complexity of an image are produced.

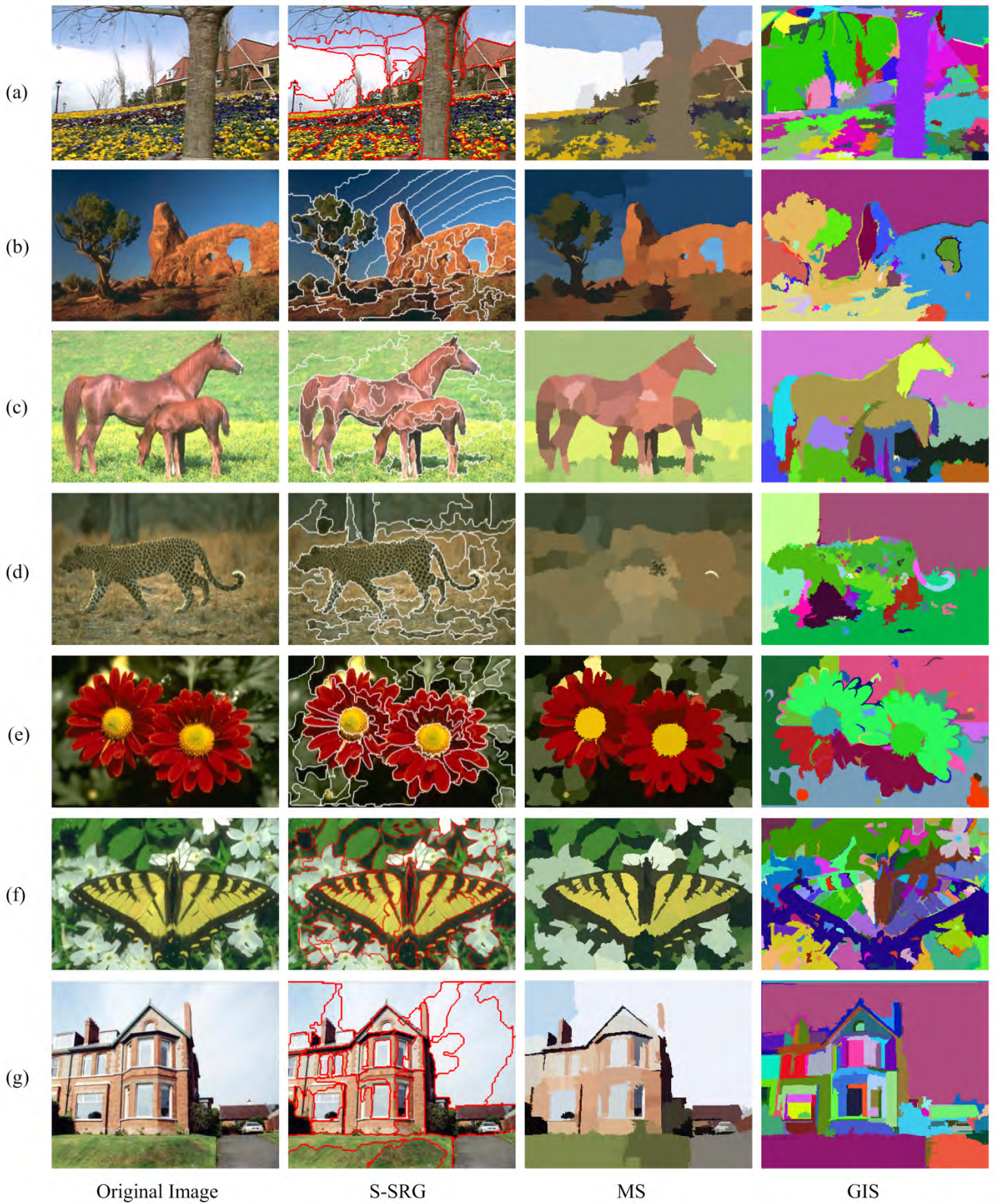

Fig. 8. The original images and results of segmentation: from left to right, the original images, and the results of our method (S-SRG), the mean shift (MS), and the graph-based image segmentation (GIS) for (a) flower garden, (b) rock, (c) horses, (d) leopard, (e) flowers, (f) butterfly, and (g) building.

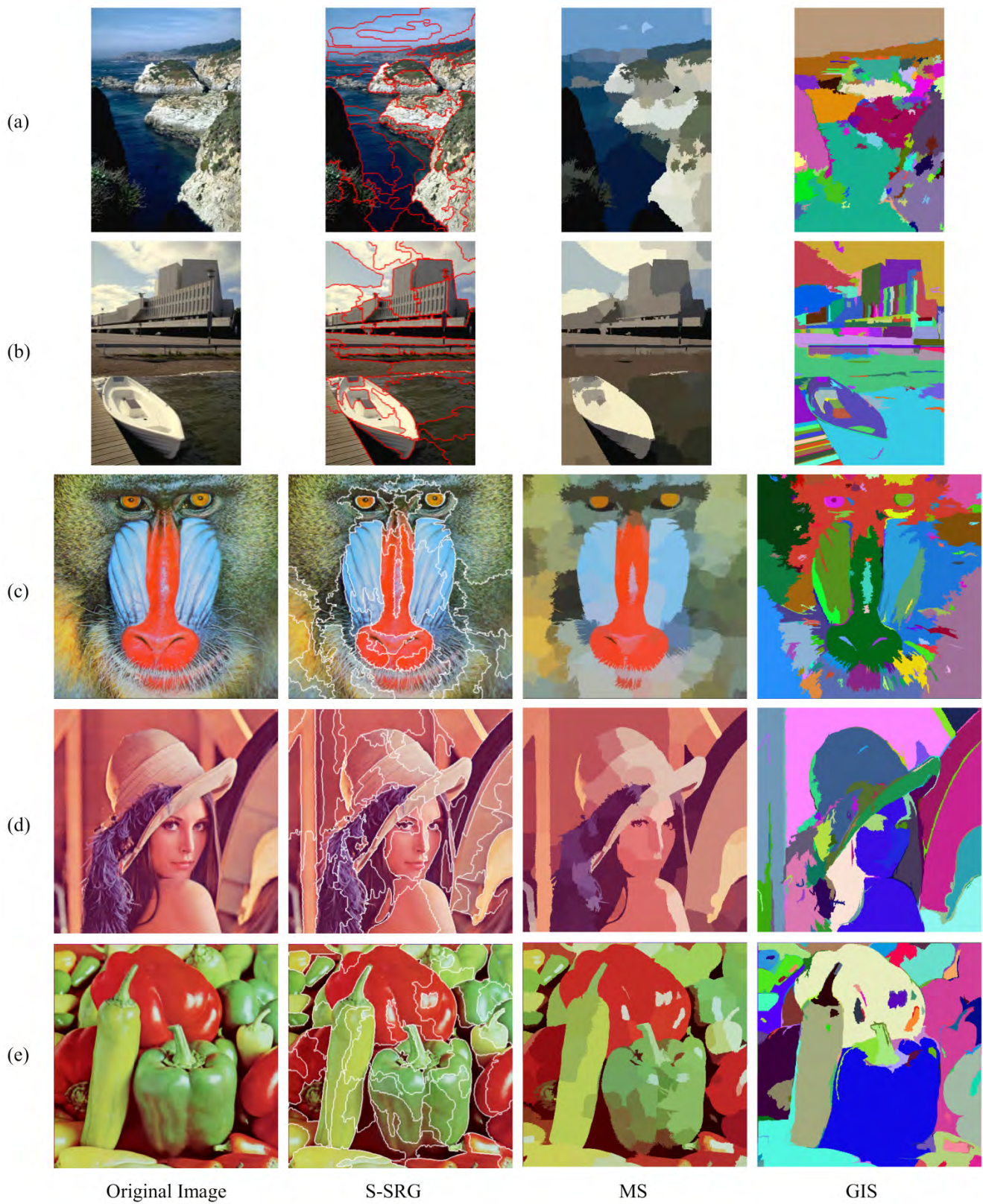

Fig. 9. The original images and results of segmentation: from left to right, the original images, and the results of our method (S-SRG), the mean shift (MS), and the graph-based image segmentation (GIS) for (a) coast, (b) boat, (c) baboon, (d) Lenna, and (b) peppers.

TABLE 2. The Total Number of Regions ($N_F$) in the Final Segmentation Results of Our Method (S-SRG), the Mean Shift (MS), and the Graph Based Image Segmentation (GIS)

| Image | S-SRG | MS | GIS |
|---|---|---|---|
| flower garden | 23 | 103 | 96 |
| rock | 43 | 117 | 54 |
| horses | 43 | 97 | 56 |
| leopard | 42 | 75 | 67 |
| flowers | 52 | 157 | 63 |
| butterfly | 54 | 152 | 157 |
| building | 32 | 81 | 86 |
| coast | 35 | 90 | 99 |
| boat | 36 | 88 | 131 |
| baboon | 38 | 240 | 128 |
| Lenna | 55 | 133 | 80 |
| peppers | 57 | 165 | 113 |

(3) The final segmentation results demonstrate that the seed region generation process using the *GCD* and *BLD* and the pixel-wise boundary determination process using the *PRD* work well to locate accurate region boundaries.

(4) Overall, our method exhibits stable performance for a variety of natural images including heavily textured areas, using the same parameter values. Our method produces considerably good segmentation results with a small number of regions.

(5) The segmentation results of our method are fairly comparable to, and in some cases better than, those of the MS algorithm, and considerably better than those of the GIS algorithm except in one case.

## 8.2 Quantitative Evaluation and Comparison

We quantitatively evaluated the segmentation results using the two functions (19) and (20) proposed in [33], which require no ground truth segmentation. These functions were designed to incorporate, explicitly or implicitly, the conditions of good image segmentation [2] listed in Section 1. These quantitative evaluation functions are often used [34], because ground truth segmentations are available only for a limited variety of images.

$$F'(I) = \frac{1}{10000(N \times M)} \sqrt{\sum_{A=1}^{Max}[R(A)]^{1+1/A}} \times \sum_{i=1}^{R} \frac{e_i^2}{\sqrt{A_i}}, \quad (19)$$

$$Q(I) = \frac{1}{10000(N \times M)} \sqrt{R} \times \sum_{i=1}^{R} \left[ \frac{e_i^2}{1 + \log A_i} + \left(\frac{R(A_i)}{A_i}\right)^2 \right], \quad (20)$$

where I is the segmented image, $N \times M$ the image size, $R$ the number of regions in the segmented image, $A_i$ the area or the number of pixels of the $i$th region, *Max* the area of the largest regions in the segmented image, $R(A)$ or $R(A_i)$ the number of regions having exactly area $A$ or $A_i$, and $e_i$ is defined as the sum of the Euclidean distances between the RGB color vectors of the pixels of region $i$ in the original image and the color vector (mean vector) attributed to the region $i$ of the segmented image. The smaller the values of $F'(I)$ and $Q(I)$ are, the better the segmentation result should be.

Table 3 shows the values of $F'(I)$ and $Q(I)$ for the test images. The bold values represent the best results among the three algorithms. According to these results, the quality of image segmentations by our S-SRG method and by MS are fairly comparable. Note that for the leopard image in Fig. 8d, the value of $Q(I)$ for MS is smaller than that for S-SRG despite the poorer quality of the result of MS. The $F'(I)$ and $Q(I)$ values of GIS segmentations are significantly larger those of S-SRG and MS, except in the case of the building image. This means that, in many cases, GIS cannot produce appropriate regions corresponding to the distribution of image characteristics.

The values produced by these functions correspond reasonably well to the visual quality of the segmentations evaluated based on our subject. However, according to the condition (b) of the conditions of good segmentation [2] listed in Section 1, over-segmentation having many small regions

TABLE 3. The Quantitative Evaluation Values of the Segmentation Results of Our Method (S-SRG), the Mean Shift (MS), and the Graph Based Image Segmentation (GIS)

| Image | $F'(I)$ | | | $Q(I)$ | | |
|---|---|---|---|---|---|---|
| | S-SRG | MS | GIS | S-SRG | MS | GIS |
| flower garden. | **106.82** | 136.68 | 193.01 | **784.87** | 932.18 | 1506.48 |
| rock | **32.06** | 76.52 | 246.38 | **229.06** | 847.73 | 3931.41 |
| horses | **58.59** | 73.39 | 148.48 | **442.75** | 595.07 | 1750.82 |
| leopard | **55.08** | 57.61 | 130.97 | 663.88 | **510.83** | 1824.89 |
| flowers | **40.68** | 45.51 | 163.82 | 266.59 | **246.13** | 1991.63 |
| butterfly | 70.37 | **57.63** | 140.75 | 481.49 | **351.59** | 1521.34 |
| building | 65.60 | **49.85** | 50.64 | 423.42 | 381.95 | **372.77** |
| coast | 83.50 | **79.89** | 189.65 | 703.23 | **547.28** | 1928.78 |
| boat | **41.07** | 54.47 | 100.11 | **299.05** | 452.98 | 1107.24 |
| baboon | 214.64 | **150.51** | 378.55 | 2960.16 | **922.71** | 4995.23 |
| Lenna | **40.10** | 53.26 | 209.92 | **327.52** | 555.70 | 2875.76 |
| peppers | 67.21 | **64.19** | 518.81 | 535.32 | **530.42** | 8607.90 |

should be more penalized.

Anyway, the results of the quantitative comparison show that the segmentation results of our method are fairly comparable to those of the MS algorithm, and considerably better than those of the GIS algorithm except in one case.

**8.3 Discussions**

According to [5], the time complexity for the pixel-based SRG method is $O((m + \log(n))n)$, where $n$ is the total number of pixels in an image, and $m$ is the total number of regions. For our method using SERs with side length $S$, the time complexity is $O((m + \log(z))z)$, where $z$ is the total number of SERs and $z = n / S^2$. Therefore, the time complexity for our method is considerably smaller than that of the pixel-based SRG method.

The inherent weakness of our method is that the resolution of the segmentation results is limited by the size of SERs: regions corresponding to objects smaller than a single SER are not produced.

Recently, [35] and [36] have proposed post-processing type image segmentation methods that use the segmentation results generated by conventional segmentation algorithms such as the mean shift [30] and graph-based image segmentation [31] as the initial regions. Although the segmentation results of these methods are excellent, they depend on the quality of the initial regions. Our method can be an alternative initial region generator for such methods.

In the segmentation results of our method, most salient objects in images are divided into several regions. Ideally, it is desired that one object correspond to one region. We will perform further research for developing a segmentation method for achieving this.

**9 CONCLUSIONS**

We have presented an efficient automatic color image segmentation method using a seeded region growing and merging method based on square elemental regions. Using square regions as the processing unit substantially reduces the time complexity of the algorithm and makes the performance stable. The experimental results show the following: our method works well and produces good segmentation results with a small number of regions; the results of our method are fairly comparable to, and in some respects better than, those of existing algorithms. Our method can be an alternative initial region generator for post-processing type image segmentation methods.

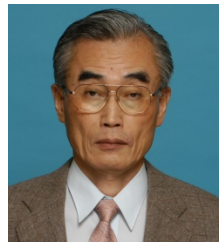
**Hisashi Shimodaira** received the BE, ME and DE degrees from Tokyo Metropolitan University in 1969, 1971, and 1982, respectively. He had been professor in the Information and Communications Department of Bunkyo University at Chigasaki City in Japan and retired from it in 2012. His research interests include computer vision and artificial intelligence.